\documentclass[sigconf]{acmart}

\renewcommand\footnotetextcopyrightpermission[1]{}
\usepackage{booktabs}
\usepackage{bbding}
\usepackage{gensymb}
\usepackage{multirow}
\usepackage{tikz}
\usepackage{comment}
\usepackage{paralist}
\usepackage{color}
\usepackage{pifont}
\usepackage{makecell}
\usepackage{caption}
\usepackage{pgfplots}
\usetikzlibrary{patterns}
\pgfplotsset{compat=1.18} %

\AtBeginDocument{%
  }

\copyrightyear{2026}
\acmYear{2026}
\setcopyright{cc}
\setcctype{by}
\acmConference[MM '26]{Proceedings of the 34th ACM International Conference on Multimedia}{November 10--14, 2026}{Rio de Janeiro, Brazil}
\acmBooktitle{Proceedings of the 34th ACM International Conference on Multimedia (MM '26), November 10--14, 2026, Rio de Janeiro, Brazil}
\acmDOI{10.1145/3767308.3836054}
\acmISBN{979-8-4007-2213-4/2026/11}

\makeatletter
\def\doclicense@image#1{%
  \IfFileExists{#1.pdf}{#1}{#1}%
}
\makeatother

\begin{document}

\title{Label-Free Target-Domain Adaptation for Unconstrained Event-Image Feature Matching via Dual-Stage Distillation}

\author{Zhonghua Yi}
\affiliation{%
  \institution{Zhejiang University}
  \city{Hangzhou}
  \country{China}}
\email{yizhonghua@zju.edu.cn}
\orcid{0009-0006-0068-7457}

\author{Hao Shi}
\affiliation{%
  \institution{Ant Group}
  \city{Beijing}
  \country{China}}
\affiliation{%
  \institution{Zhejiang University}
  \city{Hangzhou}
  \country{China}}
\email{sh531467@antgroup.com}
\orcid{0000-0003-0184-2245}

\author{Qi Jiang}
\affiliation{%
  \institution{Zhejiang University}
  \city{Hangzhou}
  \country{China}}
\email{qijiang@zju.edu.cn}
\orcid{0000-0002-7484-3030}

\author{Yufan Zhang}
\affiliation{%
  \institution{National University of Defense Technology}
  \city{Changsha}
  \country{China}}
\email{zhangyufan@nudt.edu.cn}

\author{Kailun Yang}
\affiliation{%
  \institution{Hunan University}
  \city{Changsha}
  \country{China}}
\email{kailun.yang@hnu.edu.cn}
\orcid{0000-0002-1090-667X}

\author{Kaiwei Wang}
\correspondingauthor
\affiliation{%
  \institution{Zhejiang University}
  \city{Hangzhou}
  \country{China}}
\email{wangkaiwei@zju.edu.cn}
\orcid{0000-0002-8272-3119}

\renewcommand{\shortauthors}{Yi et al.}

\begin{abstract}
Building pixel-level correspondence between event and image data is a fundamental task for multi-sensor systems. However, existing cross-modal matching methods are largely restricted by their reliance on either matching labels or strictly aligned hardware, which limits them to unlabeled and unconstrained real-world scenarios where neither matching ground truth nor prior sensor relationships are available. To address this, we propose a novel two-stage training paradigm. First, we leverage large-scale data to perform label-agnostic distillation pretraining, upgrading optimization objectives with distribution-based and contrastive losses to learn highly generalizable representations. Second, to tackle unlabeled and unconstrained downstream data, we introduce an epipolar-guided self-distillation framework. By utilizing consistency verification to isolate robust matches and incorporating geometric confidence derived from an external epipolar prior, our model can effectively self-evolve directly on target domains without any supervision. Furthermore, we introduce a rigorous cross-modal evaluation benchmark based on TUM-VIE, featuring physically separated cameras with distinct intrinsic parameters and resolutions. Extensive experiments demonstrate that our proposed method achieves state-of-the-art performance on both MVSEC and TUM-VIE pose estimation tasks. The source code and benchmark will be made publicly available at \url{https://github.com/ZhonghuaYi/nexus2-official}.
\end{abstract}


\maketitle

\section{Introduction}

Cross-modal feature matching~\cite{tuzcuouglu2024xoftr,ren2025minima,he2025matchanything} has recently attracted significant research interest, focusing on establishing pixel-level correspondences between multi-modal data. Among these tasks, event-image feature matching~\cite{yi2025ei} is particularly challenging. Due to the unique nature of event cameras, their data is represented as asynchronous point clouds in 3D spatiotemporal space, making it significantly more difficult to match with standard 2D images.

\begin{figure*}
    \centering
    \includegraphics[width=\linewidth]{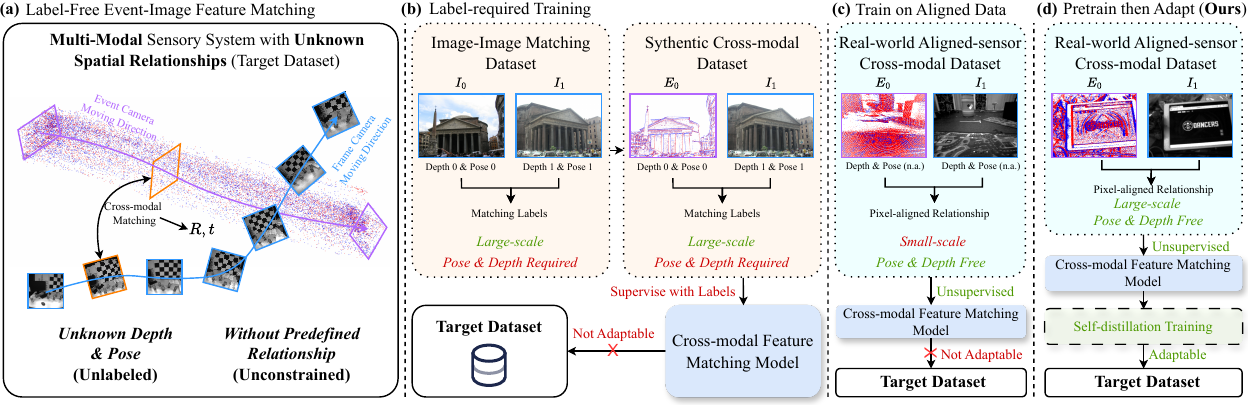}
    \vskip -1\baselineskip plus -1fil
    \caption{(a) In unconstrained downstream tasks, unknown spatial relationships and absent geometric priors prevent obtaining matching labels. (b) Supervised synthesis paradigms rely on simulated data and existing matching labels, rendering them incapable of label-free fine-tuning. (c) Alignment-dependent approaches fully rely on pixel-aligned data, failing to adapt to target datasets where such hardware relations are unavailable. (d) Our framework first performs label-agnostic pretraining for robust zero-shot features. Driven by self-distillation, the model could self-evolve directly on the unlabeled target domain.}
    \vskip -1\baselineskip plus -1fil
    \label{fig:teaser}
\end{figure*}

Unlabeled and unconstrained multi-sensor systems~\cite{zuniga2019automatic} are ubiquitous in real-world applications, since they bypass the need for expensive tracking equipment, especially in distributed sensor networks where relative sensor poses are inherently dynamic and unfixed. As illustrated in Fig.~\ref{fig:teaser}(a), our target scenario focuses on these physically decoupled, heterogeneous sensors with unknown spatial relationships. 
In such environments, the model must learn to establish correspondences on target data without access to any ground-truth matching labels through pose and depth (unlabeled), or predefined sensor relationships (unconstrained).

However, previous methods struggle to generalize to such setups and generally fall into two categories. The first category, shown in Fig.~\ref{fig:teaser}(b), relies on large-scale end-to-end training using synthetic multimodal data with matching labels. 
These methods require image datasets with known camera poses and pixel-wise depth to synthesize large amounts of multimodal data, thereby generalizing to event-image matching~\cite{ren2025minima}. 
The second category, such as EI-Nexus~\cite{yi2025ei} (Fig.~\ref{fig:teaser}(c)), relaxes the need for synthetic labels by distilling knowledge from aligned event-image pairs. However, these alignment-dependent approaches are fundamentally bottlenecked by their reliance on specialized hardware (\textit{e.g.}, coaxial DAVIS cameras) to provide pixel-perfect spatial correspondence. Consequently, they cannot adapt to target downstream datasets where such strict hardware alignment is unavailable.
To bridge this gap, we propose a two-stage training paradigm (Fig.~\ref{fig:teaser}(d)) that enables model optimization on unlabeled and unconstrained event-image data.
In the pretraining stage, while sharing a similar distillation philosophy with EI-Nexus, we significantly enhance the model's foundational robustness. Specifically, we scale up the training to a massive corpus of real-world data and upgrade the distillation objectives from simple pixel regression to a local score distribution loss and a contrastive descriptor loss. This ensures the model learns highly generalizable cross-modal features rather than overfitting to specific spatial configurations.

Our most significant departure from prior work lies in the second stage: epipolar-guided self-distillation. Unlike EI-Nexus, which is restricted to its pretraining environment, our framework allows the model to continuously self-evolve directly on unaligned target domains. We employ a teacher-student framework initialized with the pretrained model. We feed homography-warped, challenging cross-modal pairs into the student network while providing the original, unwarped event-image pairs to the teacher network. Then we perform consistency verification between the teacher's and student's matching predictions to extract high-consistency matches. Furthermore, by incorporating an epipolar geometric prior, we calculate geometric confidences based on the teacher's matches. This confidence score guides the self-distillation process, encouraging the network to focus more on matching predictions that strictly conform to the underlying epipolar geometry. This self-distillation strategy significantly boosts the pretrained model's performance on downstream data, entirely without the need for matching labels.

Furthermore, existing evaluation datasets for event-image matching suffer from notable limitations. They are typically based on DAVIS cameras~\cite{zhu2018multivehicle,mueggler2017event}, where events and frames share identical camera intrinsics, resolutions, and optical axes, or they inappropriately evaluate 2D homography estimation on 3D scenes with parallax~\cite{ren2025minima}, violating basic geometric constraints. Consequently, these datasets fail to accurately reflect feature matching performance in genuine multi-sensor systems. To address this critical gap, we construct a novel and rigorous evaluation benchmark based on the TUM-VIE dataset~\cite{klenk2021tum}. Unlike previous setups, it features spatially offset event and frame cameras with heterogeneous properties (\textit{e.g.}, distinct intrinsics and resolutions), serving as a realistic platform for unlabeled and unconstrained multi-sensor applications.

Extensive experiments on both traditional aligned and physically decoupled datasets demonstrate that our framework achieves state-of-the-art performance across MVSEC~\cite{yi2025ei} and TUM-VIE pose estimation benchmarks. 
Notably, our label-agnostic pretrained model exhibits superior zero-shot generalization, outperforming supervised baselines on the MVSEC dataset. Furthermore, our self-distillation effectively bridges severe domain gaps in unlabeled and unconstrained systems and yields a significant performance boost, achieving up to $21.9\%$ improvement on MVSEC and $16.6\%$ on the TUM-VIE benchmark compared to our zero-shot foundation.

In summary, our key contributions are as follows:
\begin{compactitem}
    \item We propose a novel two-stage training paradigm for event-image feature matching, enabling robust model deployment on unconstrained and unlabeled downstream data.
    \item We introduce an epipolar-guided self-distillation framework that leverages external geometric priors to enable the network's self-evolution without matching labels.
    \item We construct a rigorous cross-modal pose estimation benchmark based on the TUM-VIE dataset, featuring physically decoupled sensors with heterogeneous properties.
    \item Our method achieves state-of-the-art performance on both MVSEC and TUM-VIE datasets, while maintaining a lightweight architecture.
\end{compactitem}
\section{Related Work}

\subsection{Image Feature Matching}

Visual feature matching is a fundamental cornerstone of computer vision. Over the past few years, the field has witnessed a paradigm shift from handcrafted features~\cite{lowe2004distinctive,rublee2011orb} to deep-learning-based architectures. Detector-based methods~\cite{gleize2023silk,zhao2023aliked, potje2024xfeat}, such as SuperPoint~\cite{detone2018superpoint}, extract sparse keypoints and establish robust correspondences via Mutual Nearest Neighbor (MNN) or Graph Neural Networks~\cite{sarlin2020superglue,lindenberger2023lightglue,jiang2024omniglue}. Concurrently, semi-dense matching paradigms~\cite{sun2021loftr,wang2024efficient,wang2022matchformer}, estimate sub-pixel matches through a coarse-to-fine scheme. 
Dense matching methods~\cite{edstedt2023dkm,edstedt2024roma,wang2024dust3r,leroy2024grounding} leverage transformers and dense correlation volumes to achieve remarkable accuracy in texture-less and illumination-variant RGB scenes. 
However, these single-modal paradigms are inherently restricted to image-to-image matching and fail to bridge the profound domain gap between asynchronous events and static frames.

\subsection{Cross-Modal Feature Matching} 
Cross-modal feature matching~\cite{wu2025mapglue,ye20253mos,he2025matchanything} has long been an active research area, primarily focusing on establishing reliable correspondences between images captured by different sensory modalities, such as RGB-IR~\cite{tuzcuouglu2024xoftr,deng2022redfeat} pairs. 
Investigating the cross-modal link between events and images is a relatively recent endeavor. The primary obstacle is that asynchronous events lack the structural regularity of standard frames, presenting a formidable challenge. 

Early event-tracking approaches often rely on strictly aligned event-image data streams~\cite{alzugaray2020haste,gehrig2020eklt,messikommer2023data}. These methods typically detect keypoints on an initial intensity frame and subsequently track them continuously using event data with high temporal resolution~\cite{hidalgo2022event,mueggler2017event}. 
However, these are not genuine cross-modal feature matching algorithms. 
Since their operational paradigm fundamentally requires pre-existing, pixel-perfect spatial alignment between events and images to even initialize the tracking process, they do not actually solve the correspondence problem across unknown spatial gaps. Consequently, they are inapplicable to physically decoupled multi-sensor systems with unknown spatial relationships.

To explicitly address unconstrained multi-modal matching, recent learning-based architectures have emerged. Notably, MINIMA~\cite{ren2025minima} tackles this by employing a multi-modal synthesis engine to generate diverse modality representations from the MegaDepth dataset~\cite{li2018megadepth}. 
By coupling this synthesized data with the original ground-truth correspondence labels, it successfully achieves end-to-end supervised training for cross-modal matching. 
MatchAnything~\cite{he2025matchanything} also follows this pipeline.
Furthermore, EI-Nexus~\cite{yi2025ei} leverages the pre-established pixel-wise correspondences inherently provided by strictly aligned event-image pairs. Through knowledge distillation, it aligns an event-based keypoint detector with a pretrained image-based detector, thereby realizing effective event-image feature matching. 
In contrast to these approaches, our work initiates with the integration of generalizable pretraining on large-scale datasets, followed by self-supervised refinement on unlabeled and unconstrained data predicated on the pretrained model, thereby enabling the effective exploitation of downstream data.

\subsection{Self-supervised Correspondence}
Since pixel-level correspondences are difficult to obtain in the real world, self-supervised methods have been an important part of correspondence estimation.
RIPE~\cite{kunzel2025ripe} utilizes reinforcement learning driven by epipolar geometry rewards on unlabeled image pairs to extract robust keypoints without depth priors.
Sonata~\cite{wu2025sonata} employs self-distillation and spatial masking to prevent networks from relying on low-level geometric shortcuts.
In the Tracking-Any-Point (TAP) task, BootsTAP~\cite{doersch2024bootstap} refines long-term trajectories using a self-supervised student-teacher framework based on spatial transformation equivariance. 
GMRW~\cite{shrivastava2024self} formulates tracking on a space-time graph, training a global matching transformer via contrastive random walks and cycle consistency. TAPNext~\cite{zholus2025tapnext} treats tracking as a causal, self-supervised sequential masked token decoding task, whereas LEAP-Track~\cite{zhao2026learning} improves global matching efficiency using curriculum-based sparse attention.
While these methods primarily target single-modal or tracking scenarios, we extend the self-supervised paradigm to cross-modal matching by leveraging epipolar geometry to guide the self-distillation process, effectively resolving the spatial misalignment between events and images without external supervision.

\section{Method}

\subsection{Problem Formulation}
\label{sec:problem}

\noindent\textbf{Cross-Modal Feature Matching.} 
Given a pair of observations $(X_A, X_B)$ from distinct modalities $\mathcal{M}_A$ and $\mathcal{M}_B$ capturing the same 3D scene, the goal is to establish a set of pixel-level correspondences $\mathcal{C} = \{(p_A^i, p_B^j)\}$. 
Feature extraction networks $f_A, f_B$ predict keypoint locations $P \in \mathbb{R}^{N \times 2}$ and $d$-dimensional descriptors $D \in \mathbb{R}^{N \times d}$ for both modalities. A correspondence is valid if $p_A^i$ and $p_B^j$ are projections of the same 3D point, typically determined by maximizing the similarity $s(D_A^i, D_B^j)$ in a shared latent descriptor space.

\noindent\textbf{Event-Image Matching.} 
We instantiate the modalities as the event domain $\mathcal{E}$ and image domain $\mathcal{I}$. Due to hardware heterogeneity, the event $E$ and image $I$ often possess distinct resolutions and intrinsic parameters.
The task is to learn an event feature extractor $f_\theta$ such that predicted event descriptors $D_E$ and image descriptors $D_I$ are modality-invariant. For any matching pair $(p_E, p_I)$, the objective is:
\begin{equation}
    s(f_\theta(E), g_\phi(I)) \rightarrow 1,
\end{equation}
where $s(\cdot, \cdot)$ is a similarity metric (\textit{e.g.}, cosine similarity).

\noindent\textbf{Unlabeled and Unconstrained Matching Challenge.} 
Traditional optimization of $\theta$ heavily relies on either explicit ground-truth correspondences $\mathcal{C}^*$ (derived from depth and pose $\mathbf{T}$) or strict hardware-level spatial alignment provided by the dataset. 
However, in real-world distributed or mobile platforms, sensors are often physically decoupled and unconstrained, rendering $\mathcal{C}^*$ and aligned data unavailable. 
Therefore, our objective is to formulate an optimization strategy for $\theta$ using only \textbf{unlabeled, spatially unconstrained} event-image pairs, leveraging implicit geometric consistency to bridge the massive modality gap without external supervision.

\subsection{Method Overview}
\label{sec:overview}

Directly optimizing an event feature extractor $f_\theta$ on unconstrained, unlabeled data is hard due to the massive modality gap and lack of spatial alignment. To address this, we propose a two-stage bootstrapping framework that first establishes a robust baseline and subsequently drives it to self-evolve on target downstream data.

\noindent\textbf{Stage 1: Label-Agnostic pretraining (Sec.~\ref{sec:pretrain}).} 
The goal of this stage is to acquire a highly generalizable initialization $\theta_{pre}$. Inspired by EI-Nexus~\cite{yi2025ei}, we utilize aligned event-image data to guide the initial cross-modal learning. However, unlike EI-Nexus, which fundamentally assumes aligned hardware is available during target deployment, we strictly confine the use of aligned data to this pretraining phase on a large-scale dataset. Using a frozen image feature extractor $g_\phi$ as a teacher, we distill knowledge into $f_\theta$ via our proposed local score distribution and contrastive descriptor losses. Unlike methods relying on synthetic events, our pretraining utilizes real-world DAVIS recordings, ensuring the baseline captures authentic event dynamics and sensor noise, which is essential for robust zero-shot generalization.

\noindent\textbf{Stage 2: Epipolar-Guided Self-Distillation (Sec.~\ref{sec:finetune}).} 
Given the initialization $\theta_{pre}$, this stage adapts the network to specific unconstrained and unlabeled downstream scenarios without matching labels. We employ a teacher-student self-distillation scheme where the student receives homography-augmented inputs. To establish reliable pseudo-labels, we perform cross-modal consistency checks between teacher and student predictions. Furthermore, to prevent confirmation bias and error accumulation, we introduce an \textit{epipolar geometric prior}. By estimating the fundamental matrix from initial matches, we compute geometric confidence scores that dynamically weight the self-distillation loss. This mechanism acts as a strict geometric filter, forcing the model to adhere to 3D physical constraints during self-evolution in unconstrained environments.

\subsection{Label-Agnostic pretraining via Distillation}
\label{sec:pretrain}

\begin{figure}
    \centering
    \includegraphics[width=1.0\linewidth]{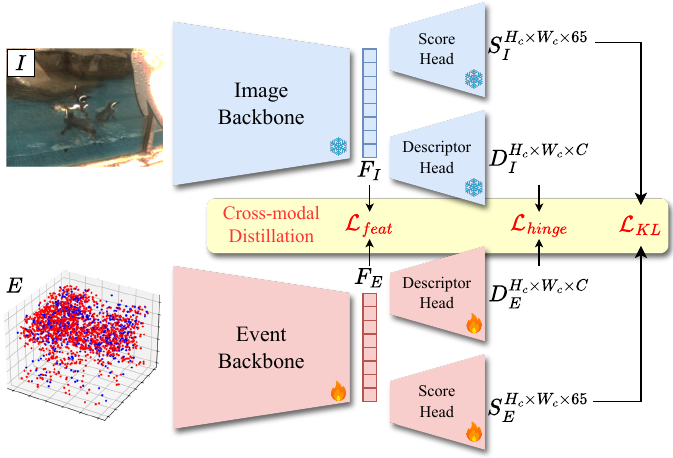}
    \vskip -1\baselineskip plus -1fil
    \caption{Overview of the pretraining. Knowledge is transferred from the frozen image teacher to the event student.}
    \label{fig:pretrain}
    \vskip -1\baselineskip plus -1fil
\end{figure}

The first stage aims to learn a robust, hardware-agnostic event representation via cross-modal distillation on large-scale aligned data (Fig.~\ref{fig:pretrain}).

\noindent\textbf{Data Selection and Quality Filtering.} 
We utilize COESOT~\cite{tang2025revisiting}, which provides large-scale, strictly aligned event-image pairs captured by DAVIS346 cameras. To prevent the student network from learning corrupted pseudo-labels caused by motion blur or HDR artifacts in standard images, we employ a Laplacian variance-based filtering strategy. Pairs with a variance below a specific threshold are excluded to ensure the structural integrity of the sharp edges required for distillation.

\noindent\textbf{Distillation Architecture.} 
We adopt the teacher-student paradigm using a frozen SuperPoint~\cite{detone2018superpoint} as the image teacher $g_\phi$. The student event network $f_\theta$ is designed with a backbone and two heads (score and descriptor) that match the teacher's output dimensions: a $H_c \times W_c \times 65$ local score distribution and a dense descriptor map. This structural alignment enables seamless knowledge transfer from the image domain to the event domain.

\noindent\textbf{Distillation Objectives.} 
Instead of naive pixel-wise regression in EI-Nexus~\cite{yi2025ei}, which limits generalization, we introduce three localized distillation losses:

\textit{1) Latent Feature Loss ($\mathcal{L}_{feat}$):} To align intermediate semantic representations, we apply an $L_2$ distance between the event and image backbones:
\begin{equation}
    \mathcal{L}_{feat} = \|F_E - F_I\|_2^2.
\end{equation}

\textit{2) Local Score Distribution Loss ($\mathcal{L}_{score}$):} To enhance keypoint robustness, we minimize the Kullback-Leibler (KL) divergence between the $8 \times 8$ local probability distributions $S_I$ and $S_E$:
\begin{equation}
    \mathcal{L}_{KL} = \frac{1}{H_c W_c} \sum_{h,w} D_{KL} ( S_I^{(h,w)} \parallel S_E^{(h,w)} ).
\end{equation}
This forces the student to learn relative keypoint likelihoods within local neighborhoods rather than absolute pixel values.

\textit{3) Contrastive Descriptor Loss ($\mathcal{L}_{desc}$):} We employ a contrastive hinge loss to establish a modality-invariant descriptor space. For a positive pair $(d_E, d_I)$ and its corresponding hard negative $d_I^-$, the loss is:
\begin{equation}
    \mathcal{L}_{hinge} = 1 - s(d_E, d_I) + \max(0, s(d_E, d_I^-) - th_{neg}),
\end{equation}
where $s(\cdot, \cdot)$ is cosine similarity. This objective explicitly aligns positive cross-modal pairs ($s \rightarrow 1$) while repelling hard negatives beyond a margin $th_{neg}$, ensuring high discriminative power.

The total pretraining loss is $\mathcal{L}_{pre} = \mathcal{L}_{feat} + \mathcal{L}_{KL} + \mathcal{L}_{hinge}$.

\begin{figure*}[!t]
    \centering
    \includegraphics[width=1.0\linewidth]{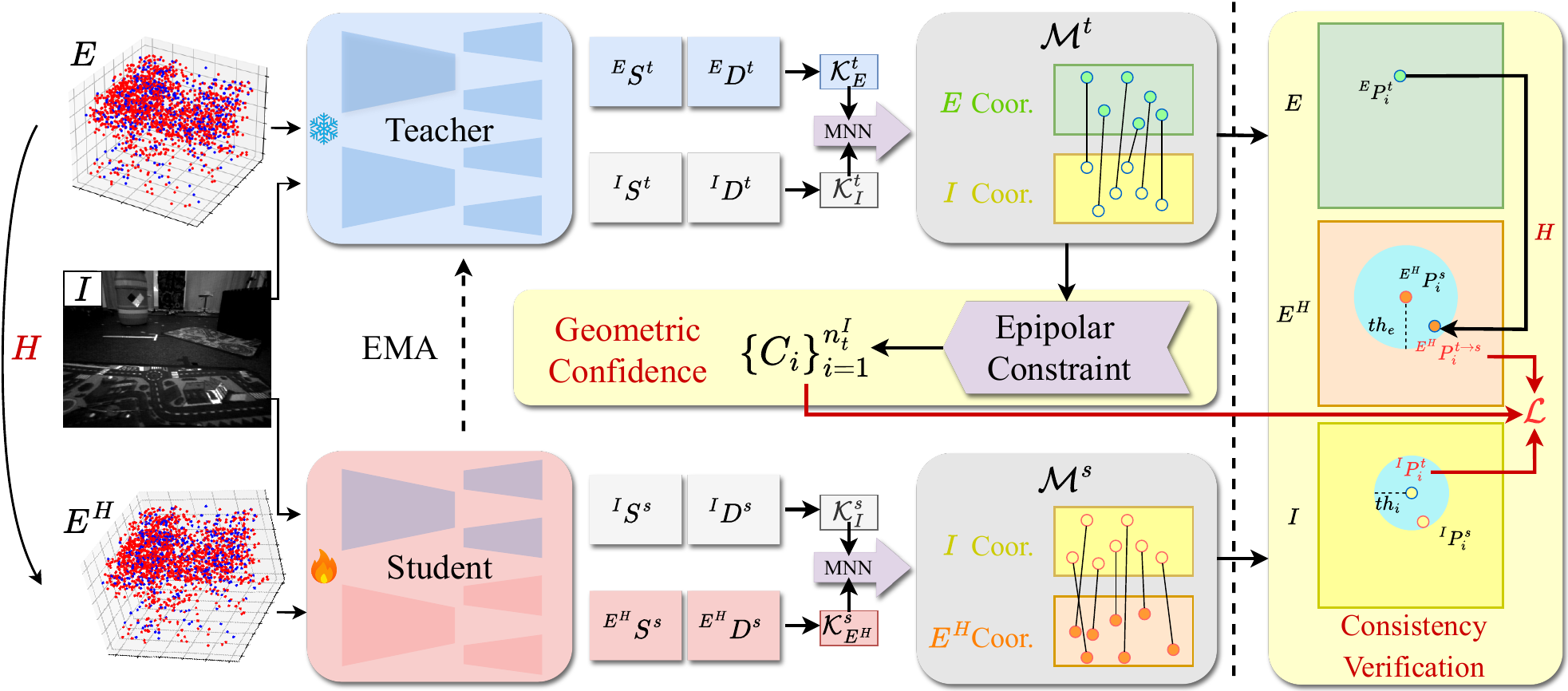}
    \vskip -1\baselineskip plus -1fil
    \caption{Epipolar-guided self-distillation procedure. The pretrained model is copied as the teacher and student network, while the teacher is updated through EMA. Student receives homography-augmented event inputs. The predicted matches from Student and Teacher are further verified by checking consistency. The Teacher's prediction was further calculated using the epipolar constraint to obtain geometric confidence, guiding the self-distillation process.}
    \vskip -1\baselineskip plus -1fil
    \label{fig:self-train}
\end{figure*}

\subsection{Epipolar-Guided Self-Distillation}
\label{sec:finetune}

To adapt the sub-optimal baseline model to downstream scenarios, where multi-modal data is completely unconstrained and unlabeled, we propose an epipolar-guided self-distillation framework. Our core insight is to drive the network to self-evolve by extracting high-quality pseudo ground-truth matches through asymmetric augmentations and geometric consistency verification. The whole self-distillation framework is shown in Fig.~\ref{fig:self-train}.

\noindent\textbf{Framework and Notation.} 
We adopt a teacher-student architecture. Let the original image coordinate system be $I$, and the original event coordinate system be $E$. 
To construct challenging asymmetric inputs, we apply a random homography $\mathbf{H}$ to the event stream $E$, yielding a warped stream $E^H = \{(\mathbf{p}_k^H, t_k, p_k)\}$, where $\mathbf{p}_k^H$ is the warped event coordinates, calculated via perspective projection: $\mathbf{p}_k^H = \text{proj}(\mathbf{H}\mathbf{p}_k)$.

Crucially, the image branches for both the teacher and the student are initialized with the SuperPoint weights and kept strictly frozen ($g_\phi$), ensuring the image feature space serves as a highly stable and consistent anchor. Let $\theta_t$ denote the parameters of the teacher's event branch, which are updated via Exponential Moving Average (EMA)~\cite{tarvainen2017mean} from the student parameters $\theta$. 
The forward passes for the teacher network, processing the unwarped original data, are formulated as:
\begin{equation}
    ({}^{I}S^t, {}^{I}D^t) = g_\phi(I),~
    ({}^{E}S^t, {}^{E}D^t) = f_{\theta_t}(E).
\end{equation}
For the student network, its image branch processes the same original image to yield outputs $({}^{I}S^s, {}^{I}D^s)$. However, its event branch processes the augmented event stream to generate predictions strictly within the transformed coordinate system $E^H$:
\begin{equation}
    ({}^{E^H}S^s, {}^{E^H}D^s) = f_\theta(E^H).
\end{equation}
By extracting sparse keypoints from the score maps and interpolating their corresponding descriptors from the descriptor maps, we obtain the complete sets of image and event keypoints. For the teacher network, we define these extracted sets as $\mathcal{K}_I^t$ and $\mathcal{K}_E^t$:
\begin{equation}
    \mathcal{K}_I^t = \left\{ \left( {}^{I}P_{i}^{t}, {}^{I}S_{i}^{t}, {}^{I}d_{i}^{t} \right) \right\}_{i=1}^{n_t^I}, \quad
    \mathcal{K}_E^t = \left\{ \left( {}^{E}P_{i}^{t}, {}^{E}S_{i}^{t}, {}^{E}d_{i}^{t} \right) \right\}_{i=1}^{n_t^E}.
\end{equation}
These two sets are then fed into a parameter-free Mutual Nearest Neighbor (MNN) matcher. We formulate this matching process as a function that yields the teacher's initial cross-modal match set $\mathcal{M}^t$:
\begin{equation}
    \mathcal{M}^t = \text{MNN}(\mathcal{K}_E^t, \mathcal{K}_I^t) = \left\{ \left( {}^{E}P_{k}^{t}, {}^{E}S_{k}^{t}, {}^{E}d_{k}^{t}, {}^{I}P_{k}^{t}, {}^{I}S_{k}^{t}, {}^{I}d_{k}^{t} \right) \right\}_{k=1}^{m_t}.
\end{equation}
Similarly, for the student network, we obtain $\mathcal{K}_I^s$, $\mathcal{K}_{E^H}^s$ and $\mathcal{M}^s = \text{MNN}(\mathcal{K}_{E^H}^s, \mathcal{K}_I^s)$.

\noindent\textbf{Cross-Modal Consistency verification.} 
To construct reliable pseudo-labels, we design a dual-threshold consistency verification between the teacher and student predictions.
First, we identify identical matched image keypoints across both networks. Given the disparate inputs furnished to the student and teacher, their resultant matched image keypoints will not achieve perfect congruence. 
Hence, it is important to identify the same image keypoints within the matching predictions derived from both models. We find the nearest neighbor in the student's image keypoints for a given teacher's image point ${}^{I}P_{i}^{t}$ and require their distance to be less than a threshold $th_i$. If this spatial consistency holds, we assume the student's image point inherits the stable features of ${}^{I}P_{i}^{t}$.

Next, we map the teacher's matched event keypoint ${}^{E}P_{i}^{t}$ from the $E$ coordinate system to the student's $E^H$ coordinate system using the known homography matrix $\mathbf{H}$:
\begin{equation}
    {}^{E^H}P_{i}^{t \rightarrow s} = \mathbf{H} \cdot {}^{E}P_{i}^{t}.
\end{equation}
We then check if this projected point aligns with the student's matched event point ${}^{E^H}P_{i}^{s}$ by requiring their distance to be less than $th_e$. If this spatial consensus is met, we establish ${}^{E^H}P_{i}^{t \rightarrow s}$ as a high-quality pseudo ground-truth event matching point for the corresponding image keypoint in the student's transformed space.

\noindent\textbf{Epipolar Geometric Confidence.} 
Relying solely on network output consistency introduces the risk of error accumulation (confirmation bias) as the student evolves. To mitigate this, we introduce an external, purely geometric filter. Without relying on specific camera intrinsics, we apply RANSAC~\cite{fischler1981random} on the teacher's overall matching results to estimate a Fundamental matrix $\mathbf{F}$. For each matched pair, we calculate its epipolar distance $err$ relative to $\mathbf{F}$. 
A smaller epipolar distance indicates a higher likelihood that the match conforms to a rigid 3D scene structure. 
We convert this error into a normalized geometric confidence score $C_i = exp(-err_i) \in (0,1]$.

\noindent\textbf{Self-Supervised Distillation Objectives.} 
Finally, we calculate the cross-modal distillation loss by enforcing the student's event predictions at the pseudo-ground-truth location ${}^{E^H}P_{i}^{t \rightarrow s}$ to mimic the corresponding stable image outputs. 

Unlike the pretraining stage, the downstream unaligned data lacks a spatial alignment prior. Consequently, formulating a KL divergence over local spatial distributions is ill-posed. Instead, we apply a direct regression loss for the scores:
\begin{equation}
    \mathcal{L}_{i}^{score} = \lVert {}^{E^H}S^s(^{E^H}P_{i}^{t \rightarrow s}) - {}^{I}S_{i}^{t} \rVert_2.
\end{equation}
For the descriptors, we employ a contrastive hinge loss. 
The pseudo-ground-truth match serves as the positive pair, while the most similar non-corresponding image descriptor ${}^{I}d_{j}^{t}$ serves as the mined hard negative:
\begin{align}
    \mathcal{L}_{i}^{desc} &= \left( 1 - {}^{E^H}D^s(^{E^H}P_{i}^{t \rightarrow s}) \cdot {}^{I}d_{i}^{t} \right) \nonumber \\
    &+ \max \left( 0, {}^{E^H}D^s(^{E^H}P_{i}^{t \rightarrow s}) \cdot {}^{I}d_{j}^{t} - th_{neg} \right).
\end{align}
The total self-supervised loss is dynamically weighted by the geometric confidence, forcing the network to focus primarily on matches that adhere to the underlying epipolar geometry:
\begin{equation}
    \mathcal{L} = \sum_{i=1}^{N} C_i \Big( \mathcal{L}_{i}^{score} + \mathcal{L}_{i}^{desc} \Big).
\end{equation}

\begin{table*}[!t]
\centering
\caption{Quantitative evaluation of cross-modal relative pose estimation on the MVSEC and TUM-VIE pose estimation benchmarks. Modality denotes whether the method performs pseudo Image-to-Image (I$\leftrightarrow$I, using event-to-image conversion) or direct Event-to-Image (E$\leftrightarrow$I) matching. ``pretrain'' and ``ft.'' represent our label-agnostic pre-training and self-supervised fine-tuning, respectively. Best results are highlighted in \textbf{bold}.}
\vskip -1\baselineskip plus -1fil
\label{tab:main_results}
\setlength{\tabcolsep}{2mm}
\resizebox{\linewidth}{!}{

\begin{tabular}{lcccc|ccc|ccc}
\Xhline{2\arrayrulewidth}
\multirow{2}{*}{\textbf{Model}} & \multirow{2}{*}{\textbf{Modality}} & \multirow{2}{*}{\textbf{Data}} & \multirow{2}{*}{\makecell{\textbf{Sensor} \\ \textbf{Relation}}} & \multirow{2}{*}{\textbf{Scheme}} & \multicolumn{3}{c|}{\textbf{MVSEC AUC@}} & \multicolumn{3}{c}{\textbf{TUM-VIE AUC@}} \\ \cline{6-11} 
 & & & &  & $5^{\circ}$ & $10^{\circ}$ & $20^{\circ}$ & $5^{\circ}$ & $10^{\circ}$ & $20^{\circ}$ \\ \Xhline{2\arrayrulewidth}
SuperPoint \cite{detone2018superpoint} & I$\leftrightarrow$I & Label-Free & Unconstrained & Self-Supervised & 0.31 & 1.50 & 5.08 & 0.00 & 0.06 & 0.22 \\
LoFTR (\textit{outdoor}) \cite{sun2021loftr} & I$\leftrightarrow$I & Label-Required & Unconstrained & Supervised & 0.23 & 0.77 & 2.94 & 0.08 & 0.19 & 0.51 \\
LoFTR (\textit{indoor}) & I$\leftrightarrow$I & Label-Required & Unconstrained & Supervised & 0.00 & 0.16 & 0.47 & 0.00 & 0.00 & 0.00 \\
RoMa (\textit{outdoor}) \cite{edstedt2024roma} & I$\leftrightarrow$I & Label-Required & Unconstrained & Supervised & 0.71 & 3.34 & 8.01 & 0.92 & 4.09 & 10.88 \\
RoMa (\textit{indoor}) & I$\leftrightarrow$I & Label-Required & Unconstrained & Supervised & 0.64 & 2.19 & 6.77 & 0.14 & 1.04 & 4.87 \\ \hline
MINIMA (\textit{LG}) \cite{ren2025minima} & E$\leftrightarrow$I & Label-Required & Unconstrained & Supervised & 1.15 & 3.20 & 7.78 & 1.07 & 5.25 & 13.90 \\
MINIMA (\textit{LoFTR}) & E$\leftrightarrow$I & Label-Required & Unconstrained & Supervised & 2.19 & 6.95 & 14.47 & 0.83 & 3.50 & 9.83 \\
MINIMA (\textit{RoMa}) & E$\leftrightarrow$I & Label-Required & Unconstrained & Supervised & 1.22 & 3.98 & 9.80 & 4.38 & 13.83 & 26.10 \\
MatchAnything (\textit{ELoFTR}) \cite{he2025matchanything} & E$\leftrightarrow$I & Label-Required & Unconstrained & Supervised & 1.55 & 4.79 & 11.25 & 0.55 & 2.31 & 6.92 \\
MatchAnything (\textit{RoMa}) & E$\leftrightarrow$I & Label-Required & Unconstrained & Supervised & 1.71 & 5.29 & 12.45 & 2.68 & 10.05 & 20.99 \\\hline
EI-Nexus (\textit{MVSEC ver.}) \cite{yi2025ei} & E$\leftrightarrow$I & Label-Free & Aligned & Distillation & 3.79 & 11.98 & 24.24 & 3.72 & 12.18 & 24.66 \\
EI-Nexus (\textit{EC ver.}) & E$\leftrightarrow$I & Label-Free & Aligned & Distillation & 2.55 & 8.48 & 18.70 & 2.19 & 8.22 & 17.02 \\ \hline
Ours (\textit{pretrain}) & E$\leftrightarrow$I & Label-Free & Aligned & Distillation & 4.47 & 13.18 & 26.44 & 3.46 & 12.63 & 24.49 \\
Ours (\textit{MVSEC ft.}) & E$\leftrightarrow$I & Label-Free & Unconstrained & Self-Distillation & \textbf{5.63} & \textbf{16.70} & \textbf{32.23} & 1.98 & 7.86 & 17.15 \\
Ours (\textit{TUM-VIE ft.}) & E$\leftrightarrow$I & Label-Free & Unconstrained & Self-Distillation & 4.04 & 12.25 & 25.33 & \textbf{4.77} & \textbf{15.35} & \textbf{28.56} \\ \Xhline{2\arrayrulewidth}
\end{tabular}

}
\vskip -1\baselineskip plus -1fil
\end{table*}

\section{Experiments}

\subsection{TUM-VIE E$\leftrightarrow$I Matching Benchmark}
\label{sec:tum_vie_benchmark}

Existing event-image matching benchmarks suffer from two major limitations: (1) relying on synthetic 2D homographies~\cite{ren2025minima, gehrig2021dsec} that ignore scene depth and lack genuine 3D parallax, or (2) utilizing idealized DAVIS sensors~\cite{yi2025ei, zhu2018multivehicle} where modalities share identical intrinsics and coaxial axes, failing to reflect modern hardware heterogeneity.
To address these gaps, we introduce the \textbf{TUM-VIE Event-Image Matching Benchmark}, a rigorous testbed based on the TUM-VIE dataset~\cite{klenk2021tum}. It features a physically decoupled, heterogeneous stereo rig with two standard cameras ($1024 \times 1024$) and two high-resolution Prophesee event cameras ($1280 \times 720$). Due to spatial offsets and distinct intrinsic properties, the modalities are naturally misaligned, providing a highly realistic environment with complex 3D parallax.

We formulate the evaluation as a cross-modal relative pose (extrinsic) estimation task. We select $4$ challenging sequences recorded under two distinct calibration configurations, encompassing a total of $8$ unique cross-modal extrinsic transformations. 
By randomly sampling $200$ timestamps, we construct a suite of $800$ pose estimation tasks. 
By evaluating recovered extrinsics against precise ground-truth annotations, this benchmark definitively measures matching robustness under genuine hardware heterogeneity and varied spatial alignments.

\subsection{Experimental Setup}
\label{sec:setup}

\subsubsection{Baselines} 

To comprehensively evaluate our method, we select representative state-of-the-art baselines across three distinct paradigms:
\noindent \textit{Single-Modal Image Matching Models:} 
To assess the zero-shot generalization of standard image matchers on cross-modal tasks, we convert the asynchronous events into pseudo-images. Specifically, we map the polarity of the latest triggered event at each pixel into an RGB format. We evaluate a diverse spectrum of models: SuperPoint~\cite{detone2018superpoint} (MNN as matcher) representing detector-based methods, LoFTR~\cite{sun2021loftr} representing semi-dense matchers, and RoMa~\cite{edstedt2024roma} representing dense matchers.
    
\noindent \textit{MINIMA~\cite{ren2025minima}} and \textit{MatchAnything~\cite{he2025matchanything}} are recent general robust cross-modal matching frameworks, which rely on ground-truth matching labels for training. 
We report results using the LightGlue (LG)~\cite{lindenberger2023lightglue}, LoFTR, and RoMa variants provided by MINIMA, as well as the ELoFTR~\cite{wang2024efficient} and RoMa variants from MatchAnything.
Due to the reliance on precise matching labels, they cannot be fine-tuned or adapted on our unlabeled downstream event-image data.
    
\noindent \textit{EI-Nexus~\cite{yi2025ei}:} It focuses specifically on event-image matching by distilling knowledge from an image detector (SuperPoint) to an event detector. We evaluate its officially released models trained on the MVSEC~\cite{zhu2018multivehicle} and EC~\cite{mueggler2017event} datasets and use MNN as the matcher. 
However, EI-Nexus fundamentally assumes that the event and image streams are densely and perfectly aligned during training. Consequently, it is incapable of training on unconstrained downstream data where spatial relationships are unknown.

\subsubsection{Datasets}

\noindent\textbf{COESOT~\cite{tang2025revisiting}.} 
We utilize the COESOT dataset for pretraining. While originally designed for object tracking, it provides a massive corpus of roughly aligned event streams and intensity frames. 
By applying our Laplacian variance filter (threshold set to $100$) to discard scenes with severe blur, we extract approximately $170$k high-quality event-image pairs for our model's zero-shot cross-modal generalization.

\noindent\textbf{MVSEC~\cite{zhu2018multivehicle}.} We utilize MVSEC as one of our downstream datasets without access to matching labels, following the evaluation pipeline as EI-Nexus~\cite{yi2025ei}. It provides $346 \times 260$ aligned event-image data.

\noindent\textbf{TUM-VIE.} As introduced in Sec.~\ref{sec:tum_vie_benchmark}, it features spatially offset sensors with high-resolution disparities ($1280 \times 720$ for events \textit{vs.} $1024 \times 1024$ for images). 
We strictly reserve $4$ complex sequences exclusively for our new evaluation benchmark, while the remaining unannotated and unaligned sequences serve as the target domain data for our self-supervised fine-tuning.

\subsubsection{Implementation Details.} 
All experiments are implemented in PyTorch and conducted on an NVIDIA RTX 4090 GPU. 

\noindent\textbf{Architecture and Input.} 
To process the asynchronous events, the input point cloud is discretized into a $16$-channel voxel grid. This is concatenated with a $1$-channel binary event mask to form a $17$-channel input tensor. Our event branch is a lightweight VGG-based architecture with only $1.48M$ parameters, comprising a backbone, a score head, and a descriptor head.

In the pretraining stage, we use a frozen SuperPoint~\cite{detone2018superpoint} as the image keypoint extractor. The contrastive margin $m$ (for $th_{neg}$) is set to $0.2$. 
We optimize the student network using the Adam optimizer with a learning rate of $5 \times 10^{-4}$ for $20$ epochs. 
To ensure training efficiency and focus, supervision is applied exclusively to spatial pixels where events occurred, effectively filtering out superfluous background information.

During self-distillation, the student is initialized with pretrained weights, while the teacher is updated via EMA. 
We randomly sample pairs from distinct timestamps within a sequence to simulate unaligned downstream data. 
The Adam optimizer is employed with a learning rate of $1 \times 10^{-4}$. 
For the consistency check, we set the image-domain threshold $th_i = 1.5$ pixels and the event-domain projection threshold $th_e = 5.0$ pixels, balancing geometric strictness with pseudo-label density. 
For dataset-specific adaptation, we fine-tune on MVSEC for $5$ epochs (full resolution, EMA decay $0.999$) and on TUM-VIE for $2$ epochs ($512 \times 512$ random crops, EMA decay $0.9999$). 
Batch sizes are set to $32$ in pretraining and $8$ in self-distillation.

\begin{figure}[!t]
    \centering
    \includegraphics[width=0.8\linewidth]{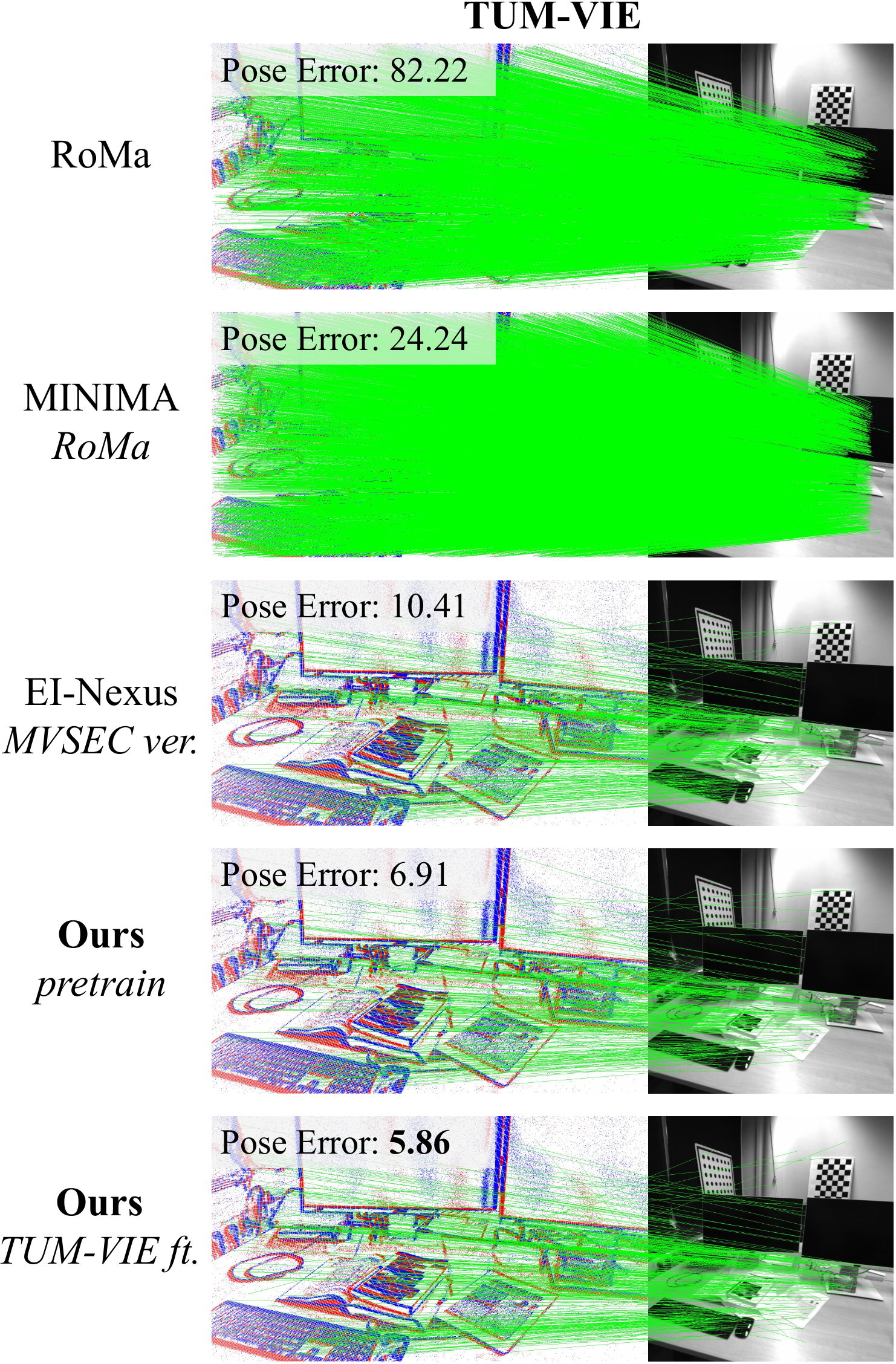}
    \vskip -1\baselineskip plus -1fil
    \caption{Qualitative results comparison on the TUM-VIE validation dataset. 
    Our pretrained model achieves better pose estimation results than previous methods without accessing the data in the target domain. 
    After self-distillation training in the target domain, our model achieves self-evolving.}
    \label{fig:vis}
    \vskip -2\baselineskip plus -1fil
\end{figure}

\subsection{Main Results}
\label{sec:main_results}
The quantitative evaluation of relative pose estimation is summarized in Tab.~\ref{tab:main_results}. 
The visualization results on TUM-VIE are shown in Fig.~\ref{fig:vis}. 
We report the Area Under the Curve (AUC) of the pose error at thresholds of $5^\circ$, $10^\circ$, and $20^\circ$. Our analysis focuses on addressing the fundamental challenges of cross-modal matching and validating the effectiveness of our proposed two-stage framework.

\noindent\textbf{Ineffectiveness of Direct Modality Translation.} 
As shown in the top section of Tab.~\ref{tab:main_results}, converting events into pseudo-images to utilize single-modal matchers (I$\leftrightarrow$I) yields critically poor performance across both datasets. 
For instance, the advanced RoMa (outdoor) model achieves only $8.01$ AUC@$20^\circ$ on MVSEC and $10.88$ on TUM-VIE. 
This demonstrates that direct event-image methods discard essential spatiotemporal dynamics, rendering standard image domain knowledge ineffective under severe modality gaps. Native cross-modal representation learning (E$\leftrightarrow$I) is strictly necessary.

\noindent\textbf{Comparison with Label-Supervised and Alignment-Dependent Baselines.} 
Our method demonstrates significant superiority by lifting restrictive training assumptions. 
First, the supervised MINIMA series suffers from severe domain overfitting due to its reliance on explicit matching labels. For example, MINIMA-RoMa’s performance collapses from $26.10$ on TUM-VIE to $9.80$ on MVSEC, revealing a critical vulnerability to distinct sensor setups. 

In addition, we compare against EI-Nexus, which relaxes label requirements but depends strictly on aligned event-image pairs for distillation. While our pretrained model already outperforms EI-Nexus on MVSEC ($26.44$ \textit{vs.} $24.24$ AUC@$20^\circ$) through more generalizable objectives, EI-Nexus's primary bottleneck is its inability to fine-tune on unaligned downstream data. 
In contrast, our self-distillation is entirely alignment-agnostic, enabling direct optimization on unconstrained target data. This adaptation successfully bridges hardware-induced domain gaps, elevating our performance to $32.23$ on MVSEC and $28.56$ on TUM-VIE. 
Ultimately, our $1.48M$ lightweight model achieves state-of-the-art accuracy without requiring matching labels or strict hardware alignment during downstream training.

\noindent\textbf{Domain Shift and the Necessity of Self-Distillation.} 
The cross-dataset evaluation results prove the effectiveness of our two-stage design paradigm. 
While our pretrained model exhibits robust zero-shot generalization, the necessity of dataset-specific adaptation becomes evident when facing severe hardware-induced domain shifts. 
For example, when our model is fine-tuned on MVSEC (\textit{MVSEC ft.}), it surges to state-of-the-art levels on MVSEC ($32.23$ AUC@$20^\circ$, $\uparrow21.9\%$ improvement), but its performance degrades when tested directly on TUM-VIE ($17.15$). 
This performance drop across disparate multi-sensor systems is a universal challenge, stemming from distinct sensor intrinsics, noise profiles, and uncalibrated spatial offsets. 
This explicitly justifies the core motivation of our Stage 2 epipolar-guided self-distillation. Because our adaptation scheme is entirely label-free and alignment-agnostic, we can seamlessly deploy and fine-tune it directly on the target unlabeled and unconstrained data. 
By doing so (\textit{TUM-VIE ft.}), the network successfully bridges this specific hardware gap, driving the AUC@$20^\circ$ from $24.49$ up to $28.56$ ($\uparrow 16.6\%$ improvement) and securing the top performance.

\begin{table}[t]
\centering

\caption{Ablation study on the impact of data filtering and various combinations of distillation objectives during pre-training. Best results are highlighted in \textbf{bold}.}
\label{tab:ablation_pretrain}
\vskip -1\baselineskip plus -1fil
\setlength{\tabcolsep}{4pt}
\resizebox{\linewidth}{!}{
\begin{tabular}{lcllc|ccc}
\Xhline{2\arrayrulewidth}
& \textbf{Data} & $L^{score}$ & $L^{desc}$ & $L_{feat}$ & \textbf{AUC@5} & \textbf{AUC@10} & \textbf{AUC@20} \\ \Xhline{2\arrayrulewidth}
Baseline & Filtered & $L_{reg}$ & $L_{reg}$ & \checkmark & 3.11 & 8.80 & 20.39 \\
1 & Filtered & $L_{KL}$ & $L_{reg}$ & \checkmark & 3.26 & 10.51 & 21.50 \\
2 & Filtered & $L_{KL}$ & $L_{hinge}$ & $\times$ & 3.29 & 11.19 & 24.40 \\ 
3 & Full & $L_{KL}$ & $L_{hinge}$ & \checkmark & 3.20 & 10.52 & 22.04 \\
4 & Filtered & $L_{KL}$ & $L_{hinge}$ & \checkmark & \textbf{4.47} & \textbf{13.18} & \textbf{26.44} \\ \Xhline{2\arrayrulewidth}
\end{tabular}
}

\end{table}

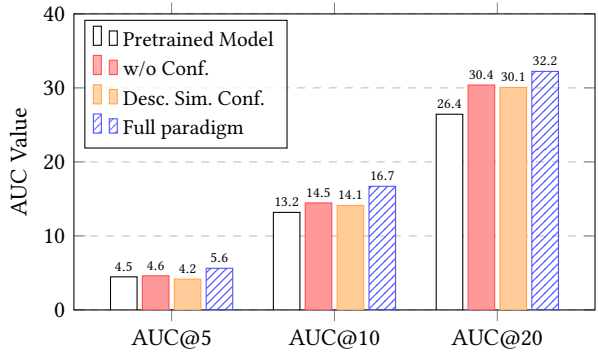
\begin{figure}[t]
\centering
\begin{tikzpicture}
\begin{axis}[
    width=1.0\linewidth,   
    height=5.5cm,
    ybar=2pt,             
    bar width=10pt,         
    enlarge x limits=0.3,  
    ymin=0, ymax=40,
    ylabel={AUC Value},
    symbolic x coords={AUC@5, AUC@10, AUC@20},
    xtick=data,
    ymajorgrids=true,
    grid style=dashed,
    nodes near coords,
    every node near coord/.append style={
        font=\tiny, 
        inner sep=2pt,
        /pgf/number format/fixed,
        /pgf/number format/precision=1 
    },
    legend style={
        at={(0.02, 0.98)},    
        anchor=north west,    
        legend columns=1,     
        font=\small,
        cells={anchor=west},  
        fill opacity=0.8,     
        text opacity=1        
    }
]

\addplot[draw=black, fill=white] 
    coordinates {(AUC@5,4.47) (AUC@10,13.18) (AUC@20,26.44)};



\addplot[fill=red!40, draw=red!70] 
    coordinates {(AUC@5,4.61) (AUC@10,14.46) (AUC@20,30.39)};

\addplot[fill=orange!40, draw=orange!70] 
    coordinates {(AUC@5,4.16) (AUC@10,14.12) (AUC@20,30.06)};

\addplot[pattern=north east lines, pattern color=blue!70, draw=blue!70] 
    coordinates {(AUC@5,5.63) (AUC@10,16.70) (AUC@20,32.23)};

\legend{
    Pretrained Model, w/o Conf., Desc. Sim. Conf., Full paradigm
}
\end{axis}
\end{tikzpicture}
\vskip -1\baselineskip plus -1fil
\caption{Ablation study on self-distillation against various variants, including distilling without confidence weighting, and weighting based purely on descriptor similarity.}
\label{fig:ablation_ss}
\vskip -1\baselineskip plus -1fil
\end{figure}

\subsection{Ablation Studies}
\label{sec:ablation}

We conduct extensive ablation studies on the MVSEC dataset to verify our architectural designs and training strategies.

\noindent\textbf{pretraining Configurations.} 
Tab.~\ref{tab:ablation_pretrain} confirms the effectiveness of our Stage 1 design. 
Using baseline (EI-Nexus) objectives achieves the lowest generalization. Upgrading from naive regression ($L_{reg}$) to distribution-based KL divergence ($L_{KL}$) and contrastive hinge loss ($L_{hinge}$) significantly boosts generalization by establishing a more discriminative cross-modal feature space.
Furthermore, data quality is critical: utilizing ``Filtered'' COESOT data instead of the ``Full'' unfiltered set increases performance from $22.04$ to $26.44$ AUC@$20^\circ$. This demonstrates that removing motion-blurred and textureless pairs is essential for stable distillation.

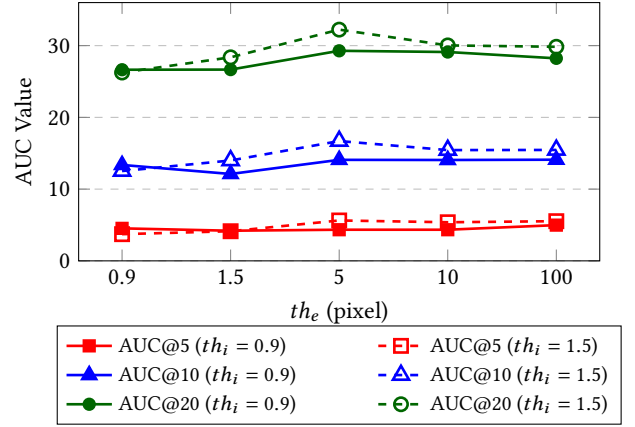
\begin{figure}[t]
\centering
\begin{tikzpicture}
\begin{axis}[
    width=1.0\linewidth,    
    height=5.cm,
    xlabel={$th_e$ (pixel)},
    ylabel={AUC Value},
    ymin=0, ymax=36,         
    symbolic x coords={0.9, 1.5, 5, 10, 100},
    xtick=data,
    ymajorgrids=true,
    grid style=dashed,
    legend style={
        at={(0.5,-0.25)},    
        anchor=north,
        legend columns=2,    
        /tikz/every even column/.append style={column sep=1cm}, 
        font=\small,
        cells={anchor=west}  
    }
]

\addplot[
    color=red,
    mark=square*,
    mark size=2pt,
    line width=1pt
] coordinates {(0.9, 4.53) (1.5, 4.20) (5, 4.34) (10, 4.34) (100, 4.98)};
\addlegendentry{AUC@5 ($th_i = 0.9$)}

\addplot[
    color=red,
    dashed,
    mark=square,
    mark options={solid}, 
    mark size=2.5pt,
    line width=1pt
] coordinates {(0.9, 3.72) (1.5, 4.13) (5, 5.63) (10, 5.38) (100, 5.53)};
\addlegendentry{AUC@5 ($th_i = 1.5$)}

\addplot[
    color=blue,
    mark=triangle*,
    mark size=3pt,
    line width=1pt
] coordinates {(0.9, 13.37) (1.5, 12.11) (5, 14.08) (10, 14.06) (100, 14.10)};
\addlegendentry{AUC@10 ($th_i = 0.9$)}

\addplot[
    color=blue,
    dashed,
    mark=triangle,
    mark options={solid},
    mark size=3.5pt,
    line width=1pt
] coordinates {(0.9, 12.52) (1.5, 13.99) (5, 16.70) (10, 15.44) (100, 15.46)};
\addlegendentry{AUC@10 ($th_i = 1.5$)}

\addplot[
    color=black!60!green, 
    mark=*,               
    mark size=2pt,
    line width=1pt
] coordinates {(0.9, 26.64) (1.5, 26.66) (5, 29.28) (10, 29.11) (100, 28.23)};
\addlegendentry{AUC@20 ($th_i = 0.9$)}

\addplot[
    color=black!60!green,
    dashed,
    mark=o,               
    mark options={solid},
    mark size=2.5pt,
    line width=1pt
] coordinates {(0.9, 26.25) (1.5, 28.37) (5, 32.23) (10, 30.03) (100, 29.84)};
\addlegendentry{AUC@20 ($th_i = 1.5$)}

\end{axis}
\end{tikzpicture}
\caption{Threshold sensitivity during self-distillation.}
\label{fig:threshold_ablation}
\vskip -1\baselineskip plus -1fil
\end{figure}

\noindent\textbf{Self-Distillation Strategies and Confidence Metrics.} 
We evaluate three settings in Stage 2 (Fig.~\ref{fig:ablation_ss}) to verify our design: (1) \textit{w/o Conf.}, self-distillation without any confidence weighting; (2) \textit{Desc. Sim. Conf.}, weighting labels via teacher-predicted descriptor similarity; and (3) \textit{Full paradigm}, our final method with epipolar-guided confidence.
Results show that naive distillation (\textit{w/o Conf.}) improves over the pretrained baseline, demonstrating the effectiveness of our base self-distillation strategy. 
In addition, relying on the teacher's own descriptor similarity (\textit{Desc. Sim. Conf.}) severely degrades performance ($30.1$ AUC@$20^\circ$). This reveals a critical circular dependency: using the network's internal features to validate its own predictions reinforces confirmation bias, making it unable to rectify incorrect matches. In contrast, our \textit{Full paradigm} achieves the highest accuracy of $32.2$ AUC@$20^\circ$, by utilizing epipolar geometry as an objective metric.

\noindent\textbf{Sensitivity of Thresholds.} 
Fig.~\ref{fig:threshold_ablation} analyzes the dual thresholds in Stage 2. For the image-domain ($th_i$), $1.5$ pixels outperforms a strict $0.9$ setting, proving that accommodating 8-connected neighborhood quantization errors is vital for robust pseudo-labeling. For the event-domain ($th_e$), we observe a fundamental trade-off between pseudo-label quality and quantity. Stringent thresholds ($th_e < 5$) starve the student network of supervisory signals, while overly relaxed settings ($th_e > 5$) introduce geometric noise. The performance peaks at $5.0$, which optimally balances geometric strictness with the volume of valid training data.
\section{Conclusion}

In this paper, we presented a novel two-stage training paradigm for event-to-image feature matching, specifically designed for unconstrained and unlabeled multi-sensor systems. By combining label-agnostic pretraining with epipolar-guided self-distillation, our framework effectively eliminates the long-standing dependence on expensive matching labels and strict hardware alignment. To provide a more realistic evaluation, we established a rigorous benchmark based on the TUM-VIE dataset, featuring genuine 3D parallax and heterogeneous sensor properties. Extensive experiments demonstrate that our lightweight model not only exhibits strong zero-shot generalization but also successfully self-evolves on unlabeled target domains, achieving state-of-the-art performance across MVSEC and TUM-VIE pose estimation tasks.

Despite the promising results, our current approach follows a detector-based paradigm, which relies on sparse keypoint extraction and matching. In the future, we aim to extend this framework to dense matching architectures. Exploring more effective dense cross-modal correspondence will be a key direction to further enhance matching robustness and coverage, particularly in scenarios with repetitive patterns or sparse textures.

\clearpage
\begin{acks}
This research was funded by the Natural Science Foundation of Zhejiang Province (Grant No. LZ24F050003), the National Natural Science Foundation of China (Grant No. 62473139), the Hunan Provincial Research and Development Project (Grant No. 2025QK3019), and the opening project of the State Key Laboratory of Autonomous Intelligent Unmanned Systems (Grant No. ZZKF2025-2-10).
\end{acks}
\bibliographystyle{ACM-Reference-Format}
\bibliography{ref}

\section{Details of TUM-VIE E$\leftrightarrow$I Matching Benchmark}
To evaluate cross-modal feature matching in physically decoupled and heterogeneous sensor systems, we construct a rigorous benchmark based on the TUM-VIE dataset~\cite{klenk2021tum}. This benchmark provides a realistic experimental platform characterized by spatially offset event and frame cameras with distinct intrinsics, resolutions, and genuine 3D parallax.

\noindent\textbf{Benchmark Structure and Splits.} 
The benchmark follows the two calibration statuses, \textbf{A} and \textbf{B}, provided by the original TUM-VIE dataset, which represent different extrinsic configurations between the event and image sensors. We utilize these predefined statuses to evaluate the model's adaptation to varying hardware setups. 
As summarized in Table~\ref{tab:supp_tumvie_splits}, we carefully partition the sequences into training and testing splits for each calibration status:
\begin{itemize}
    \item \textbf{Calibration A:} We utilize $5$ sequences for training, including \textit{mocap-desk} and the \textit{loop-floor} series. 
    The test split consists of \textit{skate-easy} and \textit{mocap-desk2}.
    \item \textbf{Calibration B:} This status includes a larger pool of $12$ training sequences, such as the \textit{office-maze}, \textit{running}, and \textit{bike} series. We reserve \textit{skate-hard} and \textit{bike-hard} for evaluation.
\end{itemize}

\noindent\textbf{Evaluation Protocol.} 
By leveraging these diverse sequences across different calibration states, our benchmark assesses the model's ability to establish correspondences under unconstrained spatial offsets and unlabeled data. The inclusion of various motion patterns (\textit{e.g.}, \textit{mocap}, \textit{skate}, and \textit{bike}) and illumination conditions (\textit{e.g.}, \textit{dark} sequences) ensures a comprehensive evaluation of feature robustness. 
This setup allows for a rigorous comparison of self-supervised and label-free matching algorithms in genuine multi-sensor environments.

\section{Architecture Details of the Event Branch}
\label{supp:event_arch}

The event branch is designed as a compact VGG-style encoder tailored for sparse event data, with $1.48M$ parameters in total. It aims to extract discriminative features that are spatially aligned with the image branch while maintaining computational efficiency.

\subsection{Input Representation}
Each event stream is first converted into a voxelized spatiotemporal representation. In our default configuration, we utilize $16$ temporal bins. To explicitly guide the network toward regions with valid observations, we append a binary event-support mask as an additional input channel. This results in a $17$-channel input tensor $\mathbf{V} \in \mathbb{R}^{17 \times H \times W}$. For the pretraining stage, we operate on a single temporal slice to ensure efficient cross-modal alignment.

\subsection{Backbone Architecture}
The backbone is responsible for extracting shared hierarchical features from the input event voxel and mask. As detailed in Table~\ref{tab:detailed_event_arch}, it consists of four convolutional stages that progressively reduce the spatial resolution to $1/8$ of the original input.

\noindent \textbf{Stages 1--2:} These stages utilize $64$ filters with $3 \times 3$ kernels. To rapidly expand the receptive field and reduce computational redundancy, each stage is followed by a $2 \times 2$ max-pooling layer, decreasing the resolution to $H/2$ and $H/4$, respectively.

\noindent \textbf{Stage 3:} This stage increases the channel dimensionality to $128$. A final $2 \times 2$ max-pooling layer is applied, resulting in a condensed feature map at $1/8$ of the input resolution.

\noindent \textbf{Stage 4:} This stage serves as the final feature integrator, consisting of two $3 \times 3$ convolutions and a $1 \times 1$ convolution, all with $128$ output channels. It maintains the $H/8 \times W/8$ resolution to produce the latent representation $\mathbf{F}$.

\subsection{Prediction Heads}
Two task-specific heads are attached to the refined feature map $\mathbf{F}$ to produce keypoint detections and descriptors.

\begin{table}[t]
\centering

\caption{Train and test splits of the TUM-VIE event-image matching benchmark. We sample two sequences for test under each calibration status.}
\label{tab:supp_tumvie_splits}
\setlength{\tabcolsep}{4pt}
\resizebox{\linewidth}{!}{
\begin{tabular}{lll}
\toprule
\textbf{Calibration} & \multicolumn{1}{c}{\textbf{Train Split}} & \multicolumn{1}{c}{\textbf{Test Split}} \\
\midrule
\textbf{A} &
\begin{minipage}[t]{4cm}
\begin{itemize}
    \item mocap-desk
    \item loop-floor0
    \item loop-floor1
    \item loop-floor2
    \item loop-floor3
\end{itemize}
\end{minipage}
&
\begin{minipage}[t]{3cm}
\begin{itemize}
    \item skate-easy
    \item mocap-desk2
\end{itemize}
\end{minipage}
\\[2ex]
\midrule
\textbf{B} &
\begin{minipage}[t]{4cm}
\begin{itemize}
    \item mocap-1d-trans
    \item mocap-3d-trans
    \item mocap-6dof
    \item mocap-shake
    \item mocap-shake2
    \item office-maze
    \item running-easy
    \item running-hard
    \item floor2-dark
    \item slide
    \item bike-easy
    \item bike-dark
\end{itemize}
\end{minipage}
&
\begin{minipage}[t]{3cm}
\begin{itemize}
    \item skate-hard
    \item bike-hard
\end{itemize}
\end{minipage}
\\
\bottomrule
\end{tabular}
}

\end{table}

\noindent\textbf{Score Head:} This head projects $\mathbf{F}$ into a $256$-dimensional latent space before mapping it to keypoint logits. The resulting coarse logits map is in $ \mathbb{R}^{65 \times H/8 \times W/8}$. 
Subsequently, the logits map is transformed to score map $\mathbf{S} \in \mathbb{R}^{65 \times H/8 \times W/8}$ by applying a softmax transformation across the 65-dimensional vector.
Then the last ``dustbin'' element is discarded, and the resultant scores are reallocated to the original $8 \times 8$ pixel grid.
Finally, we apply the input event-support mask to the score map, ensuring that detections are strictly confined to regions with physical event activity. After Non-Maximum Suppression (NMS) and border filtering, we utilize a top-k strategy to extract keypoints.

\noindent\textbf{Descriptor Head:} This head transforms $\mathbf{F}$ into a dense descriptor map $\mathbf{D} \in \mathbb{R}^{256 \times H/8 \times W/8}$. The descriptors are $L_2$-normalized. For sparse matching, we sample descriptors at the detected keypoint locations, while the dense map is reserved for downstream cross-modal distillation and fine-tuning.

\begin{table}[h]
\centering
\caption{Detailed architectural configuration of the Event Branch. $H$ and $W$ denote the input image height and width. The architecture is divided into a shared Backbone for feature extraction and two task-specific Heads.}
\label{tab:detailed_event_arch}
\resizebox{\linewidth}{!}{
\begin{tabular}{llccc}
\toprule
\textbf{Module} & \textbf{Sub-module / Layer} & \textbf{Kernel} & \textbf{Output Channels} & \textbf{Spatial Scale} \\
\midrule
\textit{Input} & Event Voxel + Mask & - & 17 & $H \times W$ \\
\midrule
\multirow{8}{*}{\textbf{Backbone}} 
 & Stage 1: Conv & $3 \times 3$ & 64 & $H \times W$ \\
 & Stage 1: Max Pooling & $2 \times 2$ & 64 & $H/2 \times W/2$ \\
 \cmidrule(lr){2-5}
 & Stage 2: Conv & $3 \times 3$ & 64 & $H/2 \times W/2$ \\
 & Stage 2: Max Pooling & $2 \times 2$ & 64 & $H/4 \times W/4$ \\
 \cmidrule(lr){2-5}
 & Stage 3: Conv & $3 \times 3$ & 128 & $H/4 \times W/4$ \\
 & Stage 3: Max Pooling & $2 \times 2$ & 128 & $H/8 \times W/8$ \\
 \cmidrule(lr){2-5}
 & Stage 4: Conv & $3 \times 3$ & 128 & $H/8 \times W/8$ \\
 & Stage 4: Conv & $3 \times 3$ & 128 & $H/8 \times W/8$ \\
 & Stage 4: Conv & $1 \times 1$ & 128 & $H/8 \times W/8$ \\
\midrule
\multirow{2}{*}{\textbf{Score Head}} 
 & Intermediate Latent & $3 \times 3$ & 256 & $H/8 \times W/8$ \\
 & Logit Projection\textsuperscript{*} & $1 \times 1$ & 65 & $H/8 \times W/8$ \\
\midrule
\multirow{2}{*}{\textbf{Descriptor Head}} 
 & Intermediate Latent & $3 \times 3$ & 256 & $H/8 \times W/8$ \\
 & Dense Descriptor Map & $1 \times 1$ & 256 & $H/8 \times W/8$ \\
\bottomrule
\end{tabular}
}
\begin{flushleft}
\footnotesize{\textsuperscript{*} The 65 channels follow the SuperPoint~\cite{detone2018superpoint} structure (8$\times$8 grid + 1 dustbin), subsequently reshaped to $H \times W$.}
\end{flushleft}
\end{table}

\section{Training Details}

\subsection{Implementation Details for Label-Agnostic Pretraining}
\label{supp:pretraining_details}

\subsubsection{Data Processing and Augmentation}
We conduct the initial pretraining on a filtered subset of the COESOT dataset~\cite{tang2025revisiting}, retaining only high-quality samples where the event stream and intensity frames are accurately aligned without severe motion blur (filtered via Laplacian variance). For each sample, we extract events within a $100\,\text{ms}$ temporal window immediately preceding the timestamp of the corresponding image frame. The event coordinates are normalized to the image resolution and voxelized into a grid with $16$ bins. To handle the sparsity of event data, an additional binary mask channel is incorporated to indicate valid event locations.

To foster geometric robustness while maintaining cross-modal consistency, we apply synchronized geometric augmentations to both modalities. These include random cropping to a $256 \times 256$ resolution and independent horizontal and vertical flips with a probability of $0.5$. Consistent with our goal of learning fundamental structural representations, no photometric distortions or event-specific point perturbations are applied during this stage.

\subsubsection{Architecture and Loss Functions}
The model adopts a two-branch architecture. The image branch is initialized with pretrained weights (\textit{e.g.}, SuperPoint~\cite{detone2018superpoint}) and remains strictly frozen to provide a stable inductive bias. The event branch is optimized to align its output space with the image branch. Both branches extract sparse keypoints and $256$-dimensional local descriptors, retaining up to $1024$ keypoints per modality for matching.
The pretraining is supervised by a composite loss function as described in the main text.

\subsubsection{Optimization Hyperparameters}
We optimize the network for $20$ epochs using the AdamW optimizer. 
The specific hyperparameter configurations are detailed in Table~\ref{tab:hyperparams}. 
The model's cross-dataset generalization is monitored by evaluating the pose estimation performance on the MVSEC dataset~\cite{zhu2018multivehicle} after each epoch.

\begin{table}[t]
\centering
\caption{Hyperparameters for label-agnostic pretraining.}
\label{tab:hyperparams}
\begin{tabular}{lr}
\toprule
\textbf{Hyperparameter} & \textbf{Value} \\
\midrule
Optimizer & AdamW \\
Base Learning Rate & $5 \times 10^{-4}$ \\
Minimum Learning Rate & $1 \times 10^{-7}$ \\
Learning Rate Schedule & Cosine Annealing \\
Weight Decay & $1 \times 10^{-5}$ \\
Betas ($\beta_1, \beta_2$) & (0.9, 0.999) \\
Epsilon ($\epsilon$) & $1 \times 10^{-8}$ \\
Effective Batch Size & 32 \\
Training Epochs & 20 \\
Max Keypoints ($N$) & 1024 \\
Descriptor Dimension & 256 \\
\bottomrule
\end{tabular}
\end{table}

\section{Implementation Details for Self-Distillation}
\label{supp:distill_details}

The self-supervised training protocol employs a teacher-student architecture. The student is actively optimized via gradient descent, while the teacher acts as a gradient-free target network updated by an Exponential Moving Average (EMA). To ensure a stable start, EMA updates begin only after a warm-up period of $1000$ iterations.

\subsection{Continuous-Space Homography Augmentation}
\label{supp:homography_details}

To cultivate modality-invariant features, we apply an asymmetric, continuous-space homography augmentation exclusively to the student's event stream, leaving the image unperturbed as a fixed geometric anchor. The planar homography is generated from a centered patch (ratio $0.8$) incorporating perspective distortion (amplitude $0.2$), isotropic scaling ($0.1$), and in-plane rotation (up to $\pm 1.0$ rad). To ensure physically plausible perturbations and avoid resampling artifacts, we enforce a strict field-of-view constraint and apply the transformation directly to the raw, asynchronous event coordinates rather than rasterized frames. Finally, the teacher's matching predictions are dynamically warped via the forward homography $\mathbf{H}$ to generate pseudo-labels directly within the student's augmented coordinate space, formulating the self-supervision as a rigorous, pixel-perfect consistency evaluation.

\subsection{Consistency Filtering and Objectives}
Because the downstream data lacks ground-truth labels, self-supervision is derived entirely from teacher-student agreement. After mapping the teacher's predictions into the student's coordinate frame via $\mathbf{H}$, we enforce a rigorous multi-step filtering pipeline to extract high-confidence pseudo-labels:
\begin{enumerate}
    \item \textbf{Mutual Consistency \& Mask Validity:} Matches must be mutually consistent across the image and event branches of the teacher. Additionally, the predicted event keypoints are verified against the binary event-support mask to discard detections in unobserved regions.
    \item \textbf{Spatial Distance Filtering:} We enforce spatial proximity constraints between corresponding features. The image-side distance threshold is set to $1.5$ pixels, while the event-side threshold is $5.0$ pixels.
    \item \textbf{Epipolar RANSAC Guidance:} We estimate a Fundamental matrix using RANSAC (with an inlier threshold of $1.0$ pixel and $0.999$ confidence) on the filtered matches to compute geometric confidence scores, further down-weighting outliers.
\end{enumerate}

The overall optimization objective combines an event-score regression term with a descriptor consistency loss. For the descriptor loss, we employ a positive alignment and a hardest-negative hinge constraint.

\subsection{Dataset-Specific Configurations (MVSEC vs. TUM-VIE)}
While the core framework is shared, several hyperparameters are tailored to the distinct characteristics of the target datasets, as summarized in Table~\ref{tab:stage2_hyperparams}.

\begin{table}[h]
\centering
\caption{Optimization hyperparameters and dataset-specific configurations for Self-Distillation. Differences between the MVSEC and TUM-VIE setups are explicitly highlighted.}
\label{tab:stage2_hyperparams}
\resizebox{\linewidth}{!}{
\begin{tabular}{lcc}
\toprule
\textbf{Hyperparameter} & \textbf{MVSEC Setup} & \textbf{TUM-VIE Setup} \\
\midrule
\multicolumn{3}{c}{\textit{Shared Optimization Parameters}} \\
\midrule
Optimizer & \multicolumn{2}{c}{AdamW} \\
Base Learning Rate & \multicolumn{2}{c}{$1 \times 10^{-4}$} \\
Minimum Learning Rate & \multicolumn{2}{c}{$1 \times 10^{-6}$ (Cosine Annealing)} \\
Weight Decay \& Betas ($\beta_1, \beta_2$) & \multicolumn{2}{c}{$1 \times 10^{-4}$ \& $(0.9, 0.999)$} \\
Gradient Clipping (Max Norm) & \multicolumn{2}{c}{1.0} \\
Mixed Precision & \multicolumn{2}{c}{Enabled (AMP)} \\
$th_i$ & \multicolumn{2}{c}{1.5 px} \\
$th_e$ & \multicolumn{2}{c}{5.0 px} \\
Effective Batch Size & \multicolumn{2}{c}{8} \\
EMA Warm-up Iterations & \multicolumn{2}{c}{1000} \\
\midrule
\multicolumn{3}{c}{\textit{Dataset-Specific Parameters}} \\
\midrule
Training Epochs / Total Steps & 5 / \textasciitilde 10k & 2 / \textasciitilde 9k \\
EMA Momentum & 0.999 & 0.9999 \\
Event Temporal Slice & 400 ms & 20ms \\
Training Resolution & $346\times 260$ & $512\times 512$ \\
\bottomrule
\end{tabular}
}
\end{table}

\noindent\textbf{Justification for Dataset-Specific Parameters.} 
The disparity in training epochs ($5$ for MVSEC and $2$ for TUM-VIE) is explicitly set to ensure a comparable number of total optimization steps (approximately $10\text{k}$) across different dataset sizes. Furthermore, the difference in the event temporal slice ($400\,\text{ms}$ for MVSEC versus $20\,\text{ms}$ for TUM-VIE) is adapted to the inherent event density of each dataset. Scenes in MVSEC typically exhibit lower event densities, requiring a longer accumulation window to form valid structural contours. Conversely, the high event density in TUM-VIE allows a concise $20\,\text{ms}$ slice to capture rich spatial details while effectively avoiding motion blur.

\section{Additional Results}

\subsection{Impact of Pretraining Datasets}
\label{supp:pretrain_datasets}

In the experimental setup of the main text, we explicitly designate MVSEC and TUM-VIE as the unlabeled and unconstrained downstream target datasets. To strictly evaluate the model's zero-shot generalization and self-supervised adaptation capabilities in authentic, unseen scenarios, we utilized exclusively the large-scale Filtered COESOT dataset during the pretraining phase. To further investigate the impact of pretraining data distribution and scale on the model's feature representation capabilities, we provide supplementary comparative experiments in Table~\ref{tab:supp_pretrain_datasets} using different pretraining datasets. Notably, all ablation experiments strictly employ the same loss function configurations (\textit{i.e.}, local score distribution loss and contrastive descriptor loss) and training hyperparameters as described in the main text.

Based on the experimental results in Table~\ref{tab:supp_pretrain_datasets}, we observe a critical dependency on both the diversity and scale of the pretraining corpus. When the model is pretrained exclusively on the scene-homogeneous MVSEC dataset, it achieves competitive performance on the homologous MVSEC test set (reaching an AUC@$20^\circ$ of $27.14$). However, when directly evaluated on the TUM-VIE dataset—which introduces heterogeneous sensor properties and a significant domain shift—its performance experiences a substantial drop, dropping to an AUC@$20^\circ$ of $19.66$. In stark contrast, pretraining solely on the Filtered COESOT dataset yields a strong zero-shot score of $24.49$ on TUM-VIE. This notable difference demonstrates that relying purely on scene-homogeneous data is insufficient to endow the model with cross-domain generalization capabilities, even when optimized with our proposed distillation objectives. Consequently, this corroborates the rationale and necessity of utilizing the diverse COESOT dataset as our foundational pretraining corpus in the main text. 

Beyond the critical need for scene diversity, our results demonstrate that scaling the pretraining data effectively further enhances the model's capability. To evaluate architectural scalability, we constructed a joint pretraining corpus combining Filtered COESOT with MVSEC. This multi-source dataset not only improves the AUC@$20^\circ$ on MVSEC to $28.48$ but also pushes the zero-shot generalization on the entirely unseen TUM-VIE dataset to a new high of $24.75$. This consistent performance gain underscores the efficacy of our proposed cross-modal distillation paradigm in absorbing and leveraging diverse data sources. Ultimately, it proves that by continuously scaling up aligned real-world pretraining data, the generalization capacity of the model's foundational features can be systematically and directly enhanced.

\begin{table}[!t]
\centering
\caption{\textbf{Impact of pre-training datasets.} Under the same training settings, COESOT demonstrates better generalization than MVSEC. Furthermore, our pretraining also scales the generalization performance as the input data increases.}

\label{tab:supp_pretrain_datasets}
\setlength{\tabcolsep}{2mm}
\resizebox{\linewidth}{!}{

\begin{tabular}{lcccccc}
\Xhline{2\arrayrulewidth}
\multirow{2}{*}{\textbf{Dataset}} & \multicolumn{3}{c}{\textbf{MVSEC AUC}} & \multicolumn{3}{c}{\textbf{TUM-VIE AUC}} \\ \cline{2-7} 
 & $5^{\circ}$ & $10^{\circ}$ & $20^{\circ}$ & $5^{\circ}$ & $10^{\circ}$ & $20^{\circ}$ \\ \Xhline{2\arrayrulewidth}
MVSEC & 4.47 & 13.45 & 27.14 & 2.31 & 9.45 & 19.66 \\
Filtered COESOT & 4.47 & 13.18 & 26.44 & 3.46 & 12.63 & 24.49 \\ 
Filtered COESOT + MVSEC & 5.66 & 15.44 & 28.48 & 3.49 & 12.41 & 24.75 \\\Xhline{2\arrayrulewidth}
\end{tabular}

}

\end{table}

\subsection{Synergy Between Pretraining and Self-Distillation}
\label{supp:synergy}

In Table~\ref{tab:supp_pretrain_and_self_distillation}, we further investigate the relationship between the quality of the pretrained initialization (Stage 1) and the final performance achieved after applying epipolar-guided self-distillation (Stage 2). We evaluate two distinct pretrained models—one trained exclusively on Filtered COESOT and another trained on the combined Filtered COESOT + MVSEC corpus. Both models are evaluated on the MVSEC target domain before and after the self-distillation phase.

The results reveal a compelling synergy between our pretraining and self-distillation stages. Most notably, applying the self-distillation framework yields substantial and consistent improvements regardless of the initial pretraining dataset. For instance, the baseline model pretrained solely on COESOT improves its AUC@$20^\circ$ from $26.44$ to $32.23$, demonstrating the robustness of our label-free self-evolution paradigm in bridging hardware-induced domain gaps. Beyond this consistent enhancement, the experiments demonstrate that a superior initialization unlocks a correspondingly higher optimization ceiling. When the model is initialized with stronger foundational features (such as those pretrained on the combined COESOT and MVSEC corpus), it naturally begins with a higher zero-shot baseline ($28.48$ vs. $26.44$). Furthermore, after undergoing the same self-distillation process, it achieves an even higher final performance of $32.87$ AUC@$20^\circ$. This constructive synergy indicates that our self-distillation framework does not merely saturate at a fixed performance bottleneck; instead, it actively leverages the richer geometric priors embedded in a superior pretrained model to extract higher-quality pseudo-labels, ultimately pushing the upper bound for downstream adaptation.

\begin{table}[!t]
\centering
\caption{\textbf{Synergy between pretraining and self-distillation.} We compare the relative pose estimation performance on MVSEC before ($\times$) and after (\checkmark) self-distillation using different pretrained initializations. The results indicate that our self-distillation consistently improves performance, and a stronger pretrained foundational model unlocks a higher upper bound for the final adaptation.}
\label{tab:supp_pretrain_and_self_distillation}
\setlength{\tabcolsep}{2mm}
\resizebox{\linewidth}{!}{
\begin{tabular}{lcccc}
\Xhline{2\arrayrulewidth}
\multirow{2}{*}{\textbf{Pretraining Dataset}} & \multirow{2}{*}{\textbf{Self-Distillation}} & \multicolumn{3}{c}{\textbf{MVSEC AUC}} \\ \cline{3-5} 
 & & $5^{\circ}$ & $10^{\circ}$ & $20^{\circ}$  \\ \Xhline{2\arrayrulewidth}
Filtered COESOT & $\times$ & 4.47 & 13.18 & 26.44 \\ 
Filtered COESOT & \checkmark & 5.63 & 16.70 & 32.23 \\ \hline
Filtered COESOT + MVSEC & $\times$ & 5.66 & 15.44 & 28.48 \\
Filtered COESOT + MVSEC & \checkmark & 5.86 & 17.53 & 32.87 \\ \Xhline{2\arrayrulewidth}
\end{tabular}
}
\end{table}

\subsection{Additional Qualitative Results}

\begin{figure*}
    \centering
    \includegraphics[width=1.0\linewidth]{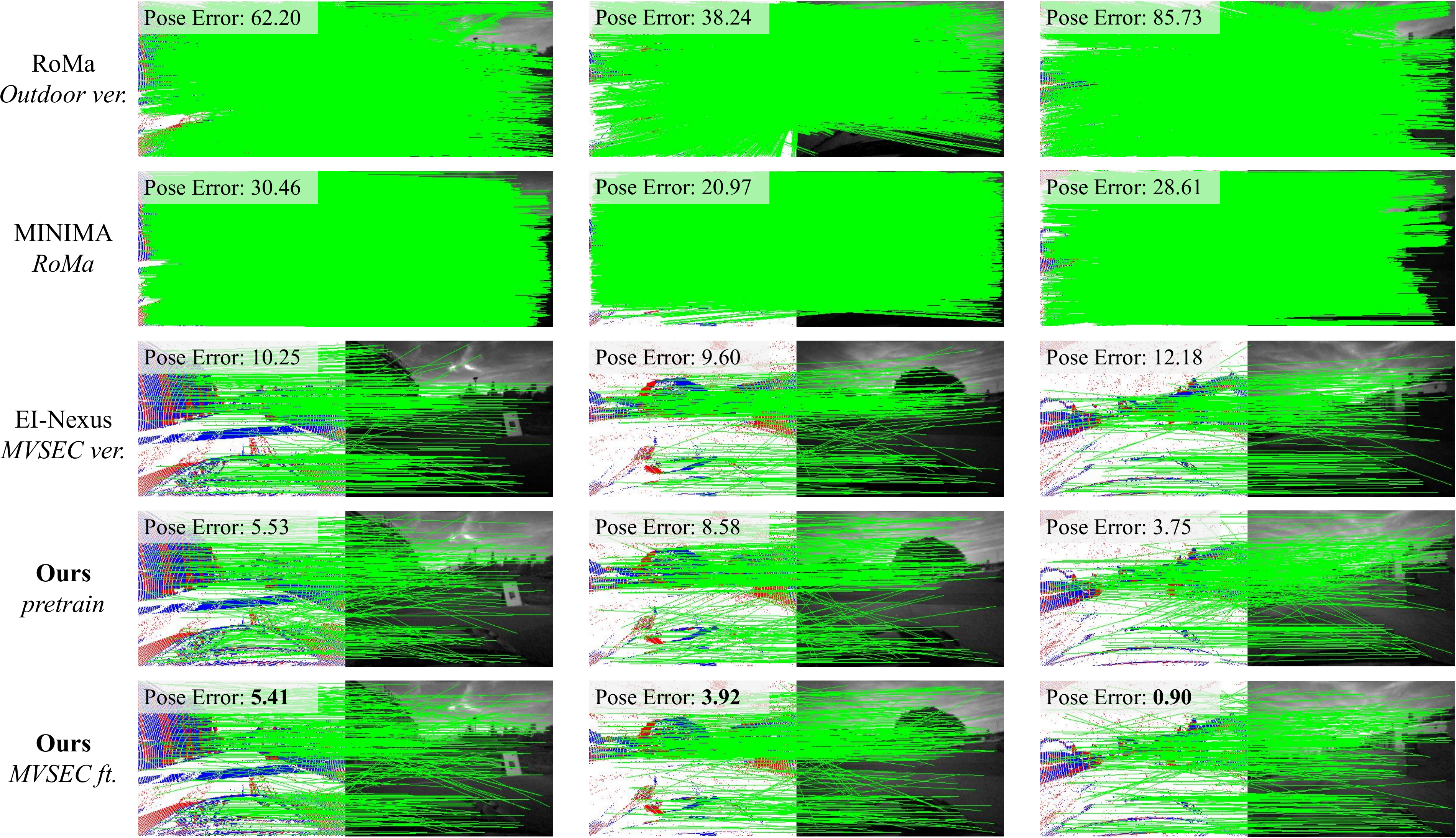}
    \vskip -1\baselineskip plus -1fil
    \caption{Qualitative results of pose estimation on the MVSEC dataset.}
    \label{fig:supp_mvsec_vis}
\end{figure*}

\begin{figure*}
    \centering
    \includegraphics[width=1.0\linewidth]{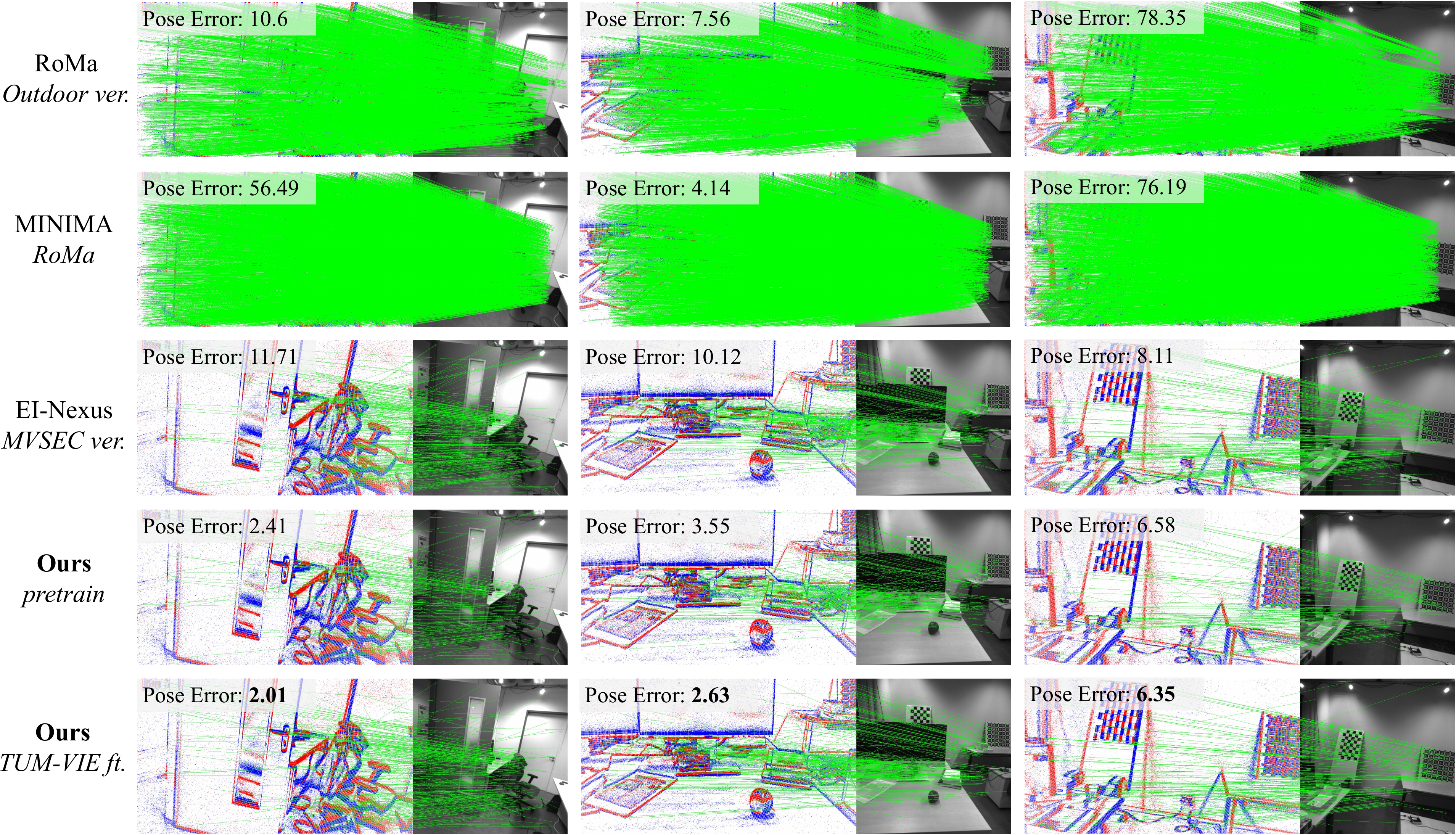}
    \vskip -1\baselineskip plus -1fil
    \caption{Qualitative results of pose estimation on the TUM-VIE dataset.}
    \label{fig:supp_tumvie_vis}
\end{figure*}

This section provides extended qualitative evaluations for cross-modal pose estimation on the MVSEC (Fig.~\ref{fig:supp_mvsec_vis}) and TUM-VIE (Fig.~\ref{fig:supp_tumvie_vis}) datasets. These visualizations further characterize the performance of our framework in unlabeled and unconstrained scenarios.

\noindent\textbf{Geometric Consistency: Sparse vs. Dense Matching.}
Visual analysis reveals a fundamental performance discrepancy between dense correspondence frameworks (\textit{e.g.}, RoMa~\cite{edstedt2024roma} and the MINIMA-variant~\cite{ren2025minima}) and sparse matching architectures. While dense matchers generate spatially continuous correspondence fields, their geometric precision often degrades in textureless or motion-blurred regions of the event stream. This leads to a high ratio of non-discriminative matches that negatively impact the robustness of RANSAC-based pose solvers. Conversely, sparse architectures leverage keypoints distilled from robust image anchors. By prioritizing feature saliency and geometric stability, these sparse models maintain a higher inlier ratio, ensuring more precise relative pose estimation under challenging cross-modal conditions.

\noindent\textbf{Comparative Analysis with Distillation-based Baselines.}
While our framework shares a distillation-based sparse matching philosophy with EI-Nexus, it offers distinct advantages in representation learning and domain adaptation. First, the integration of local score distribution and contrastive descriptor losses in the pretraining stage yields a more robust zero-shot foundation than the regression-based targets utilized in EI-Nexus. The visualizations indicate that our keypoints exhibit superior cross-modal pose estimation performance, even before target-domain fine-tuning.

Second, our framework facilitates a self-distillation phase that addresses the inherent limitations of label-required training. Unlike EI-Nexus, which is constrained by its alignment-required design, our model undergoes unsupervised optimization directly on the target domain that does not provide matching labels or aligned events and frames. By incorporating epipolar geometric constraints into the self-supervision loop, the model effectively accounts for the spatial offsets in physically decoupled systems. This results in correspondences that are more accurately aligned with the underlying 3D structure and more uniformly distributed across the scene, as evidenced by the improved pose estimation metrics in the qualitative plots.

\end{document}